# Ultra-Fine Entity Typing

**Eunsol Choi**[†]  **Omer Levy**[†]  **Yejin Choi**[†♯]  **Luke Zettlemoyer**[†]

[†]Paul G. Allen School of Computer Science & Engineering, University of Washington
[♯]Allen Institute for Artificial Intelligence, Seattle WA
{eunsol,omerlevy,yejin,lsz}@cs.washington.edu

## Abstract

We introduce a new entity typing task: given a sentence with an entity mention, the goal is to predict a set of free-form phrases (e.g. skyscraper, songwriter, or criminal) that describe appropriate types for the target entity. This formulation allows us to use a new type of distant supervision at large scale: head words, which indicate the type of the noun phrases they appear in. We show that these ultra-fine types can be crowd-sourced, and introduce new evaluation sets that are much more diverse and fine-grained than existing benchmarks. We present a model that can predict open types, and is trained using a multitask objective that pools our new head-word supervision with prior supervision from entity linking. Experimental results demonstrate that our model is effective in predicting entity types at varying granularity; it achieves state of the art performance on an existing fine-grained entity typing benchmark, and sets baselines for our newly-introduced datasets.[1]

## 1 Introduction

Entities can often be described by very fine grained types. Consider the sentences "Bill robbed John. He was arrested." The noun phrases "John," "Bill," and "he" have very specific types that can be inferred from the text. This includes the facts that "Bill" and "he" are both likely "criminal" due to the "robbing" and "arresting," while "John" is more likely a "victim" because he was "robbed." Such fine-grained types (victim, criminal) are important for context-sensitive tasks such as coreference resolution and question answering (e.g. "Who was the victim?"). Inferring such types for each mention (John, he) is not possible given current typing models that only predict relatively coarse types and only consider named entities.

[1]Our data and model can be downloaded from: http://nlp.cs.washington.edu/entity_type

| Sentence with Target Entity | Entity Types |
|---|---|
| During the Inca Empire, {the Inti Raymi} was the most important of four ceremonies celebrated in Cusco. | event, festival, **ritual, custom, ceremony, party, celebration** |
| {They} have been asked to appear in court to face the charge. | person, **accused, suspect, defendant** |
| Ban praised Rwanda's commitment to the UN and its role in {peacemaking operations}. | event, **plan, mission, action** |

Table 1: Examples of entity mentions and their annotated types, as annotated in our dataset. The entity mentions are bold faced and in the curly brackets. The bold blue types do not appear in existing fine-grained type ontologies.

To address this challenge, we present a new task: given a sentence with a target entity mention, predict free-form noun phrases that describe appropriate types for the role the target entity plays in the sentence. Table 1 shows three examples that exhibit a rich variety of types at different granularities. Our task effectively subsumes existing fine-grained named entity typing formulations due to the use of a very large type vocabulary and the fact that we predict types for all noun phrases, including named entities, nominals, and pronouns.

Incorporating fine-grained entity types has improved entity-focused downstream tasks, such as relation extraction (Yaghoobzadeh et al., 2017a), question answering (Yavuz et al., 2016), query analysis (Balog and Neumayer, 2012), and coreference resolution (Durrett and Klein, 2014). These systems used a relatively coarse type ontology. However, manually designing the ontology is a challenging task, and it is difficult to cover all pos-sible concepts even within a limited domain. This can be seen empirically in existing datasets, where the label distribution of fine-grained entity typing datasets is heavily skewed toward coarse-grained types. For instance, annotators of the OntoNotes dataset (Gillick et al., 2014) marked about half of the mentions as "other," because they could not find a suitable type in their ontology (see Figure 1 for a visualization and Section 2.2 for details).

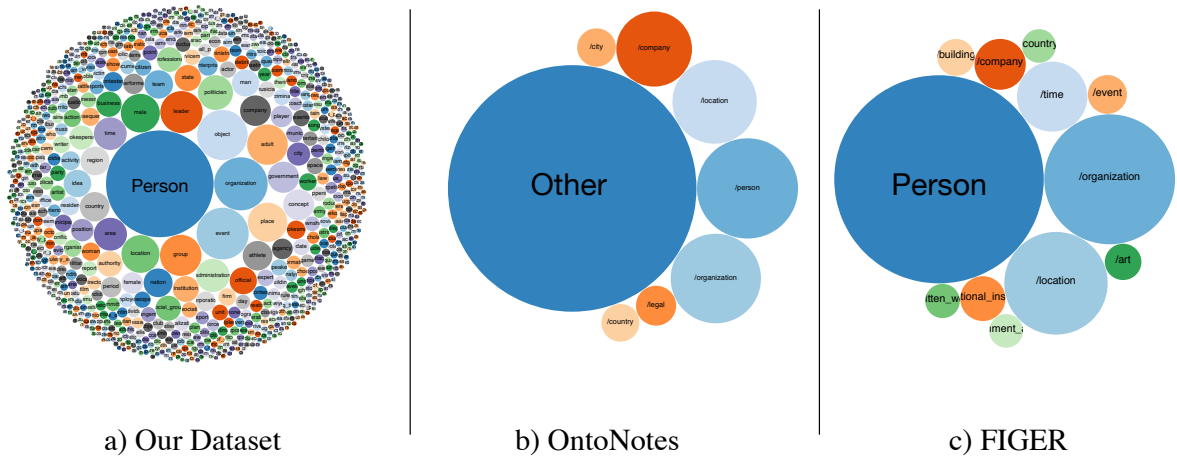

a) Our Dataset    b) OntoNotes    c) FIGER

Figure 1: A visualization of all the labels that cover 90% of the data, where a bubble's size is proportional to the label's frequency. Our dataset is much more diverse and fine grained when compared to existing datasets (OntoNotes and FIGER), in which the top 5 types cover 70-80% of the data.

Our more open, ultra-fine vocabulary, where types are free-form noun phrases, alleviates the need for hand-crafted ontologies, thereby greatly increasing overall type coverage. To better understand entity types in an unrestricted setting, we crowdsource a new dataset of 6,000 examples. Compared to previous fine-grained entity typing datasets, the label distribution in our data is substantially more *diverse* and *fine-grained*. Annotators easily generate a wide range of types and can determine with 85% agreement if a type generated by another annotator is appropriate. Our evaluation data has over 2,500 unique types, posing a challenging learning problem.

While our types are harder to predict, they also allow for a new form of contextual distant supervision. We observe that text often contains cues that explicitly match a mention to its type, in the form of the mention's head word. For example, "the incumbent chairman of the African Union" is a type of "chairman." This signal complements the supervision derived from linking entities to knowledge bases, which is context-oblivious. For example, "Clint Eastwood" can be described with dozens of types, but context-sensitive typing would prefer "director" instead of "mayor" for the sentence "Clint Eastwood won 'Best Director' for Million Dollar Baby."

We combine head-word supervision, which provides ultra-fine type labels, with traditional signals from entity linking. Although the problem is more challenging at finer granularity, we find that mixing fine and coarse-grained supervision helps significantly, and that our proposed model with a multitask objective exceeds the performance of existing entity typing models. Lastly, we show that head-word supervision can be used for previous formulations of entity typing, setting the new state-of-the-art performance on an existing fine-grained NER benchmark.

## 2 Task and Data

Given a sentence and an entity mention $e$ within it, the task is to predict a set of natural-language phrases $T$ that describe the type of $e$. The selection of $T$ is context sensitive; for example, in "Bill Gates has donated billions to eradicate malaria," Bill Gates should be typed as "philanthropist" and not "inventor." This distinction is important for context-sensitive tasks such as coreference resolution and question answering (e.g. "Which philanthropist is trying to prevent malaria?").

We annotate a dataset of about 6,000 mentions via crowdsourcing (Section 2.1), and demonstrate that using an large type vocabulary substantially increases annotation coverage and diversity over existing approaches (Section 2.2).

## 2.1 Crowdsourcing Entity Types

To capture multiple domains, we sample sentences from Gigaword (Parker et al., 2011), OntoNotes (Hovy et al., 2006), and web articles (Singh et al., 2012). We select entity mentions by taking maximal noun phrases from a constituency parser (Manning et al., 2014) and mentions from a coreference resolution system (Lee et al., 2017).

We provide the sentence and the target entity mention to five crowd workers on Mechanical Turk, and ask them to annotate the entity's type. To encourage annotators to generate fine-grained types, we require at least one general type (e.g. person, organization, location) and two specific types (e.g. doctor, fish, religious institute), from a type vocabulary of about 10K frequent noun phrases. We use WordNet (Miller, 1995) to expand these types automatically by generating all their synonyms and hypernyms based on the most common sense, and ask five different annotators to validate the generated types. Each pair of annotators agreed on 85% of the binary validation decisions (i.e. whether a type is suitable or not) and 0.47 in Fleiss's $\kappa$. To further improve consistency, the final type set contained only types selected by at least 3/5 annotators. Further crowdsourcing details are available in the supplementary material.

Our collection process focuses on precision. Thus, the final set is diverse but not comprehensive, making evaluation non-trivial (see Section 5).

## 2.2 Data Analysis

We collected about 6,000 examples. For analysis, we classified each type into three disjoint bins:

- 9 **general** types: person, location, object, organization, place, entity, object, time, event
- 121 **fine-grained** types, mapped to fine-grained entity labels from prior work (Ling and Weld, 2012; Gillick et al., 2014) (e.g. film, athlete)
- 10,201 **ultra-fine** types, encompassing every other label in the type space (e.g. detective, lawsuit, temple, weapon, composer)

On average, each example has 5 labels: 0.9 general, 0.6 fine-grained, and 3.9 ultra-fine types. Among the 10,000 ultra-fine types, 2,300 unique types were actually found in the 6,000 crowdsourced examples. Nevertheless, our distant supervision data (Section 3) provides positive training examples for every type in the entire vocabulary, and our model (Section 4) can and does predict from a 10K type vocabulary. For example, the model correctly predicts "television network" and "archipelago" for some mentions, even though that type never appears in the 6,000 crowdsourced examples.

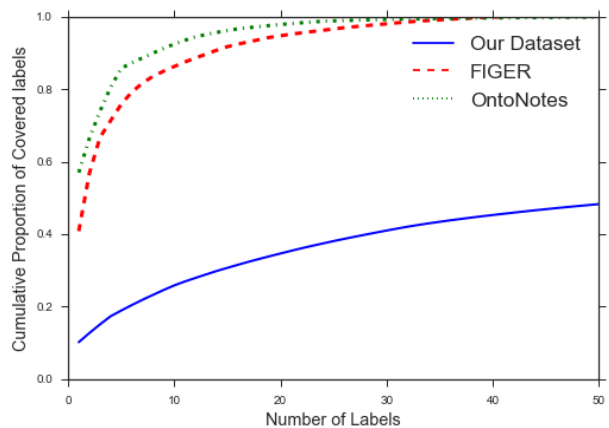

Figure 2: The label distribution across different evaluation datasets. In existing datasets, the top 4 or 7 labels cover over 80% of the labels. In ours, the top 50 labels cover less than 50% of the data.

**Improving Type Coverage** We observe that prior fine-grained entity typing datasets are heavily focused on coarse-grained types. To quantify our observation, we calculate the distribution of types in FIGER (Ling and Weld, 2012), OntoNotes (Gillick et al., 2014), and our data. For examples with multiple types ($|T| > 1$), we counted each type $1/|T|$ times.

Figure 2 shows the percentage of labels covered by the top $N$ labels in each dataset. In previous enitity typing datasets, the distribution of labels is highly skewed towards the top few labels. To cover 80% of the examples, FIGER requires only the top 7 types, while OntoNotes needs only 4; our dataset requires 429 different types.

Figure 1 takes a deeper look by visualizing the types that cover 90% of the data, demonstrating the diversity of our dataset. It is also striking that more than half of the examples in OntoNotes are classified as "other," perhaps because of the limitation of its predefined ontology.

**Improving Mention Coverage** Existing datasets focus mostly on named entity mentions, with the exception of OntoNotes, which contained nominal expressions. This has implications on the transferability of FIGER/OntoNotes-based models to tasks such as coreference resolution, which need to analyze all types of entity mentions (pronouns, nominal expressions, and named entity mentions). Our new dataset provides a well-rounded benchmark with roughly 40% pronouns, 38% nominal expressions, and 22% named entity mentions. The case of pronouns is particularly interesting, since the mention itself provides little information.

| Source | Example Sentence | Labels | Size | Prec. |
|---|---|---|---|---|
| Head Words | **Western powers that brokered the proposed deal in Vienna** are likely to balk, said Valerie Lincy, a researcher with the Wisconsin Project. | power | 20M | 80.4% |
| | Alexis Kaniaris, CEO of the organizing company Europartners, explained, speaking in a radio program in **national radio station NET**. | radio, station, radio_station | | |
| Entity Linking + Definitions | **Toyota** recalled more than 8 million vehicles globally over sticky pedals that can become entrapped in floor mats. | manufacturer | 2.7M | 77.7% |
| Entity Linking + KB | Iced Earth's musical style is influenced by many traditional heavy metal groups such as **Black Sabbath**. | person, artist, actor, author, musician | 2.5M | 77.6% |

Table 2: Distant supervision examples and statistics. We extracted the headword and Wikipedia definition supervision from Gigaword and Wikilink corpora. KB-based supervision is mapped from prior work, which used Wikipedia and news corpora.

## 3 Distant Supervision

Training data for fine-grained NER systems is typically obtained by linking entity mentions and drawing their types from knowledge bases (KBs). This approach has two limitations: recall can suffer due to KB incompleteness (West et al., 2014), and precision can suffer when the selected types do not fit the context (Ritter et al., 2011). We alleviate the recall problem by mining entity mentions that were linked to Wikipedia in HTML, and extract relevant types from their encyclopedic definitions (Section 3.1). To address the precision issue (context-insensitive labeling), we propose a new source of distant supervision: automatically extracted nominal head words from raw text (Section 3.2). Using head words as a form of distant supervision provides fine-grained information about named entities and nominal mentions. While a KB may link "the 44th president of the United States" to many types such as author, lawyer, and professor, head words provide only the type "president", which is relevant in the context.

We experiment with the new distant supervision sources as well as the traditional KB supervision. Table 2 shows examples and statistics for each source of supervision. We annotate 100 examples from each source to estimate the noise and usefulness in each signal (precision in Table 2).

### 3.1 Entity Linking

For KB supervision, we leveraged training data from prior work (Ling and Weld, 2012; Gillick et al., 2014) by manually mapping their ontology to our 10,000 noun type vocabulary, which covers 130 of our labels (general and fine-grained).[2] Section 6 defines this mapping in more detail.

To improve both entity and type coverage of KB supervision, we use definitions from Wikipedia. We follow Shnarch et al. () who observed that the first sentence of a Wikipedia article often states the entity's type via an "is a" relation; for example, "Roger Federer is a Swiss professional tennis player." Since we are using a large type vocabulary, we can now mine this typing information.[3] We extracted descriptions for 3.1M entities which contain 4,600 unique type labels such as "competition," "movement," and "village."

We bypass the challenge of automatically linking entities to Wikipedia by exploiting existing hyperlinks in web pages (Singh et al., 2012), following prior work (Ling and Weld, 2012; Yosef et al., 2012). Since our heuristic extraction of types from the definition sentence is somewhat noisy, we use a more conservative entity linking policy[4] that yields a signal with similar overall accuracy to KB-linked data.

---
[2] Data from: https://github.com/shimaokasonse/NFGEC

[3] We extract types by applying a dependency parser (Manning et al., 2014) to the definition sentence, and taking nouns that are dependents of a copular edge or connected to nouns linked to copulars via appositive or conjunctive edges.

[4] Only link if the mention contains the Wikipedia entity's name *and* the entity's name contains the mention's head.

## 3.2 Contextualized Supervision

Many nominal entity mentions include detailed type information within the mention itself. For example, when describing Titan V as "the newly-released graphics card", the head words and phrases of this mention ("graphics card" and "card") provide a somewhat noisy, but very easy to gather, context-sensitive type signal.

We extract nominal head words with a dependency parser (Manning et al., 2014) from the Gigaword corpus as well as the Wikilink dataset. To support multiword expressions, we included nouns that appear next to the head if they form a phrase in our type vocabulary. Finally, we lowercase all words and convert plural to singular.

Our analysis reveals that this signal has a comparable accuracy to the types extracted from entity linking (around 80%). Many errors are from the parser, and some errors stem from idioms and transparent heads (e.g. "parts of capital" labeled as "part"). While the headword is given as an input to the model, with heavy regularization and multitasking with other supervision sources, this supervision helps encode the context.

## 4 Model

We design a model for predicting sets of types given a mention in context. The architecture resembles the recent neural AttentiveNER model (Shimaoka et al., 2017), while improving the sentence and mention representations, and introducing a new multitask objective to handle multiple sources of supervision. The hyperparameter settings are listed in the supplementary material.

**Context Representation** Given a sentence $x_1, \ldots, x_n$, we represent each token $x_i$ using a pre-trained word embedding $w_i$. We concatenate an additional location embedding $l_i$ which indicates whether $x_i$ is before, inside, or after the mention. We then use $[x_i; l_i]$ as an input to a bidirectional LSTM, producing a contextualized representation $h_i$ for each token; this is different from the architecture of Shimaoka et al. 2017, who used two separate bidirectional LSTMs on each side of the mention. Finally, we represent the context $c$ as a weighted sum of the contextualized token representations using MLP-based attention:

$$a_i = \text{SoftMax}_i(v_a \cdot \text{relu}(W_a h_i))$$

Where $W_a$ and $v_a$ are the parameters of the attention mechanism's MLP, which allows interaction between the forward and backward directions of the LSTM before computing the weight factors.

**Mention Representation** We represent the mention $m$ as the concatenation of two items: (a) a character-based representation produced by a CNN on the entire mention span, and (b) a weighted sum of the pre-trained word embeddings in the mention span computed by attention, similar to the mention representation in a recent coreference resolution model (Lee et al., 2017). The final representation is the concatenation of the context and mention representations: $r = [c; m]$.

**Label Prediction** We learn a type label embedding matrix $W_t \in \mathbb{R}^{n \times d}$ where $n$ is the number of labels in the prediction space and $d$ is the dimension of $r$. This matrix can be seen as a combination of three sub matrices, $W_{general}, W_{fine}, W_{ultra}$, each of which contains the representations of the general, fine, and ultra-fine types respectively. We predict each type's probability via the sigmoid of its inner product with $r$: $y = \sigma(W_t r)$. We predict every type $t$ for which $y_t > 0.5$, or $\arg\max y_t$ if there is no such type.

**Multitask Objective** The distant supervision sources provide partial supervision for ultra-fine types; KBs often provide more general types, while head words usually provide only ultra-fine types, without their generalizations. In other words, the absence of a type at a different level of abstraction does not imply a negative signal; e.g. when the head word is "inventor", the model should not be discouraged to predict "person".

Prior work used a customized hinge loss (Abhishek et al., 2017) or max margin loss (Ren et al., 2016a) to improve robustness to noisy or incomplete supervision. We propose a multitask objective that reflects the characteristic of our training dataset. Instead of updating all labels for each example, we divide labels into three bins (general, fine, and ultra-fine), and update labels only in bin containing at least one positive label. Specifically, the training objective is to minimize $J$ where $t$ is the target vector at each granularity:

$$\begin{aligned} J_{\text{all}} = &J_{\text{general}} \cdot \mathbb{1}_{\text{general}}(t) \\ &+ J_{\text{fine}} \cdot \mathbb{1}_{\text{fine}}(t) \\ &+ J_{\text{ultra}} \cdot \mathbb{1}_{\text{ultra}}(t) \end{aligned}$$

Where $\mathbb{1}_{\text{category}}(t)$ is an indicator function that checks if $t$ contains a type in the category, and

| Model | Dev | | | | Test | | | |
|---|---|---|---|---|---|---|---|---|
| | MRR | P | R | F1 | MRR | P | R | F1 |
| AttentiveNER | 0.221 | **53.7** | 15.0 | 23.5 | 0.223 | **54.2** | 15.2 | 23.7 |
| Our Model | **0.229** | 48.1 | **23.2** | **31.3** | **0.234** | 47.1 | **24.2** | **32.0** |

Table 3: Performance of our model and AttentiveNER (Shimaoka et al., 2017) on the new entity typing benchmark, using same training data. We show results for both development and test sets.

| Train Data | Total | | | | General (1918) | | | Fine (1289) | | | Ultra-Fine (7594) | | |
|---|---|---|---|---|---|---|---|---|---|---|---|---|---|
| | MRR | P | R | F1 | P | R | F1 | P | R | F1 | P | R | F1 |
| All | **0.229** | 48.1 | **23.2** | **31.3** | 60.3 | 61.6 | 61.0 | 40.4 | **38.4** | **39.4** | 42.8 | 8.8 | 14.6 |
| – Crowd | 0.173 | 40.1 | 14.8 | 21.6 | 53.7 | 45.6 | 49.3 | 20.8 | 18.5 | 19.6 | 54.4 | 4.6 | 8.4 |
| – Head | 0.220 | **50.3** | 19.6 | 28.2 | 58.8 | **62.8** | 60.7 | **44.4** | 29.8 | 35.6 | **46.2** | 4.7 | 8.5 |
| – EL | 0.225 | 48.4 | 22.3 | 30.6 | **62.2** | 60.1 | **61.2** | 40.3 | 26.1 | 31.7 | 41.4 | **9.9** | **16.0** |

Table 4: Results on the development set for different type granularity and for different supervision data with our model. In each row, we remove a single source of supervision. Entity linking (EL) includes supervision from both KB and Wikipedia definitions. The numbers in the first row are example counts for each type granularity.

$J_{\text{category}}$ is the category-specific logistic regression objective:

$$J = -\sum_i t_i \cdot \log(y_i) + (1 - t_i) \cdot \log(1 - y_i)$$

## 5 Evaluation

**Experiment Setup** The crowdsourced dataset (Section 2.1) was randomly split into train, development, and test sets, each with about 2,000 examples. We use this relatively small manually-annotated training set (*Crowd* in Table 4) alongside the two distant supervision sources: entity linking (KB and Wikipedia definitions) and head words. To combine supervision sources of different magnitudes (2K crowdsourced data, 4.7M entity linking data, and 20M head words), we sample a batch of equal size from each source at each iteration. We reimplement the recent AttentiveNER model (Shimaoka et al., 2017) for reference.[5]

We report macro-averaged precision, recall, and F1, and the average mean reciprocal rank (MRR).

**Results** Table 3 shows the performance of our model and our reimplementation of AttentiveNER. Our model, which uses a multitask objective to learn finer types without punishing more general types, shows recall gains at the cost of drop in precision. The MRR score shows that our model is slightly better than the baseline at ranking correct types above incorrect ones.

Table 4 shows the performance breakdown for different type granularity and different supervision. Overall, as seen in previous work on fine-grained NER literature (Gillick et al., 2014; Ren et al., 2016a), finer labels were more challenging to predict than coarse grained labels, and this issue is exacerbated when dealing with ultra-fine types. All sources of supervision appear to be useful, with crowdsourced examples making the biggest impact. Head word supervision is particularly helpful for predicting ultra-fine labels, while entity linking improves fine label prediction. The low general type performance is partially because of nominal/pronoun mentions (e.g. "it"), and because of the large type inventory (sometimes "location" and "place" are annotated interchangeably).

**Analysis** We manually analyzed 50 examples from the development set, four of which we present in Table 5. Overall, the model was able to generate accurate general types and a diverse set of type labels. Despite our efforts to annotate a comprehensive type set, the gold labels still miss many potentially correct labels (example (a): "man" is reasonable but counted as incorrect). This makes the precision estimates lower than the actual performance level, with about half the precision errors belonging to this category. Real precision errors include predicting co-hyponyms (example (b): "accident" instead of "attack"), and types that may be true, but are not supported by the context.

---
[5] We use the AttentiveNER model with no engineered features or hierarchical label encoding (as a hierarchy is not clear in our label setting) and let it predict from the same label space, training with the same supervision data.

|     |            |                                                                                                                                                                   |
| --- | ---------- | ----------------------------------------------------------------------------------------------------------------------------------------------------------------- |
| (a) | Example    | Bruguera said {he} had problems with his left leg and had grown tired early during the match .                                                                    |
|     | Annotation | **person, athlete, player, adult, male, contestant**                                                                                                              |
|     | Prediction | **person, athlete, player, adult, male, contestant**, defendant, man                                                                                              |
| (b) | Example    | {The explosions} occurred on the night of October 7 , against the Hilton Taba and campsites used by Israelis in Ras al-Shitan.                                    |
|     | Annotation | **event** *calamity, attack, disaster*                                                                                                                            |
|     | Prediction | **event,** accident                                                                                                                                               |
| (c) | Example    | Similarly , Enterprise was considered for refit to replace Challenger after {the latter} was destroyed , but Endeavour was built from structural spares instead . |
|     | Annotation | *object, spacecraft, rocket, thing, vehicle, shuttle*                                                                                                             |
|     | Prediction | event                                                                                                                                                             |
| (d) | Context    | " There is a wealth of good news in this report , and I 'm particularly encouraged by the progress {we} are making against AIDS , " HHS Secretary Donna Shalala said in a statement. |
|     | Annotation | **government, group,** *organization,hospital,administration,socialist*                                                                                           |
|     | Prediction | **government, group**, person                                                                                                                                     |

Table 5: Example and predictions from our best model on the development set. Entity mentions are marked with curly brackets, the correct predictions are boldfaced, and the missing labels are italicized and written in red.

We found that the model often abstained from predicting any fine-grained types. Especially in challenging cases as in example (c), the model predicts only general types, explaining the low recall numbers (28% of examples belong to this category). Even when the model generated correct fine-grained types as in example (d), the recall was often fairly low since it did not generate a complete set of related fine-grained labels.

Estimating the performance of a model in an incomplete label setting and expanding label coverage are interesting areas for future work. Our task also poses a potential modeling challenge; sometimes, the model predicts two incongruous types (e.g. "location" and "person"), which points towards modeling the task as a joint set prediction task, rather than predicting labels individually. We provide sample outputs on the project website.

## 6 Improving Existing Fine-Grained NER with Better Distant Supervision

We show that our model and distant supervision can improve performance on an existing fine-grained NER task. We chose the widely-used OntoNotes (Gillick et al., 2014) dataset which includes nominal and named entity mentions.[6]

---
[6]While we were inspired by FIGER (Ling and Weld, 2012), the dataset presents technical difficulties. The test set has only 600 examples, and the development set was labeled with distant supervision, not manual annotation. We therefore focus our evaluation on OntoNotes.

**Augmenting the Training Data** The original OntoNotes training set (ONTO in Tables 6 and 7) is extracted by linking entities to a KB. We supplement this dataset with our two new sources of distant supervision: Wikipedia definition sentences (WIKI) and head word supervision (HEAD) (see Section 3). To convert the label space, we manually map a single noun from our natural-language vocabulary to each formal-language type in the OntoNotes ontology. 77% of OntoNote's types directly correspond to suitable noun labels (e.g. "doctor" to "/person/doctor"), whereas the other cases were mapped with minimal manual effort (e.g. "musician" to "person/artist/music", "politician" to "/person/political_figure"). We then expand these labels according to the ontology to include their hypernyms ("/person/political_figure" will also generate "/person"). Lastly, we create negative examples by assigning the "/other" label to examples that are not mapped to the ontology. The augmented dataset contains 2.5M/0.6M new positive/negative examples, of which 0.9M/0.1M are from Wikipedia definition sentences and 1.6M/0.5M from head words.

**Experiment Setup** We compare performance to other published results and to our reimplementation of AttentiveNER (Shimaoka et al., 2017). We also compare models trained with different sources of supervision. For this dataset, we did not use our multitask objective (Section 4), since expanding types to include their ontological hypernyms largely eliminates the partial supervision as-sumption. Following prior work, we report macro- and micro-averaged F1 score, as well as accuracy (exact set match).

|  | Acc. | Ma-F1 | Mi-F1 |
|---|---|---|---|
| AttentiveNER++ | 51.7 | 70.9 | 64.9 |
| AFET (Ren et al., 2016a) | 55.1 | 71.1 | 64.7 |
| LNR (Ren et al., 2016b) | 57.2 | 71.5 | 66.1 |
| Ours (ONTO+WIKI+HEAD) | **59.5** | **76.8** | **71.8** |

Table 6: Results on the OntoNotes fine-grained entity typing test set. The first two models (AttentiveNER++ and AFET) use only KB-based supervision. LNR uses a filtered version of the KB-based training set. Our model uses all our distant supervision sources.

| Model | Training Data | | | Performance | | |
|---|---|---|---|---|---|---|
|  | ONTO | WIKI | HEAD | Acc. | MaF1 | MiF1 |
| Attn. NER | ✓ |  |  | 46.5 | 63.3 | 58.3 |
|  | ✓ | ✓ | ✓ | 53.7 | 72.8 | 68.0 |
| Ours | ✓ |  |  | 41.7 | 64.2 | 59.5 |
|  | ✓ | ✓ |  | 48.5 | 67.6 | 63.6 |
|  | ✓ |  | ✓ | 57.9 | 73.0 | 66.9 |
|  |  | ✓ | ✓ | 60.1 | 75.0 | 68.7 |
|  | ✓ | ✓ | ✓ | **61.6** | **77.3** | **71.8** |

Table 7: Ablation study on the OntoNotes fine-grained entity typing development. The second row isolates dataset improvements, while the third row isolates the model.

**Results** Table 6 shows the overall performance on the test set. Our combination of model and training data shows a clear improvement from prior work, setting a new state-of-the art result.[7]

In Table 7, we show an ablation study. Our new supervision sources improve the performance of both the AttentiveNER model and our own. We observe that every supervision source improves performance in its own right. Particularly, the naturally-occurring head-word supervision seems to be the prime source of improvement, increasing performance by about 10% across all metrics.

**Predicting Miscellaneous Types** While analyzing the data, we observed that over half of the mentions in OntoNotes' development set were annotated only with the miscellaneous type ("/other"). For both models in our evaluation, detecting the miscellaneous category is substantially easier than producing real types (94% F1 vs. 58% F1 with our best model). We provide further details of this analysis in the supplementary material.

---
[7] We did not compare to a system from (Yogatama et al., 2015), which reports slightly higher test number (72.98 micro F1) as they used a different, unreleased test set.

## 7 Related Work

Fine-grained NER has received growing attention, and is used in many applications (Gupta et al., 2017; Ren et al., 2017; Yaghoobzadeh et al., 2017b; Raiman and Raiman, 2018). Researchers studied typing in varied contexts, including mentions in specific sentences (as we consider) (Ling and Weld, 2012; Gillick et al., 2014; Yogatama et al., 2015; Dong et al., 2015; Schutze et al., 2017), corpus-level prediction (Yaghoobzadeh and Schütze, 2016), and lexicon level (given only a noun phrase with no context) (Yao et al., 2013).

Recent work introduced fine-grained type ontologies (Rabinovich and Klein, 2017; Murty et al., 2017; Corro et al., 2015), defined using Wikipedia categories (100), Freebase types (1K) and WordNet senses (16K). However, they focus on named entities, and data has been challenging to gather, often approximating gold annotations with distant supervision. In contrast, (1) our ontology contains any frequent noun phrases that depicts a type, (2) our task goes beyond named entities, covering every noun phrase (even pronouns), and (3) we provide crowdsourced annotations which provide context-sensitive, fine grained type labels.

Contextualized fine-grained entity typing is related to selectional preference (Resnik, 1996; Pantel et al., 2007; Zapirain et al., 2013; de Cruys, 2014), where the goal is to induce semantic generalizations on the type of arguments a predicate prefers. Rather than focusing on predicates, we condition on the entire sentence to deduce the arguments' types, which allows us to capture more nuanced types. For example, not every type that fits "**He** played the violin in his room" is also suitable for "**He** played the violin in the Carnegie Hall". Entity typing here can be connected to argument finding in semantic role labeling.

To deal with noisy distant supervision for KB population and entity typing, researchers used multi-instance multi-label learning (Surdeanu et al., 2012; Yaghoobzadeh et al., 2017b) or custom losses (Abhishek et al., 2017; Ren et al., 2016a). Our multitask objective handles noisy supervision by pooling different distant supervision sources across different levels of granularity.

## 8 Conclusion

Using virtually unrestricted types allows us to expand the standard KB-based training methodology with typing information from Wikipedia definitions and naturally-occurring head-word supervision. These new forms of distant supervision boost performance on our new dataset as well as on an existing fine-grained entity typing benchmark. These results set the first performance levels for our evaluation dataset, and suggest that the data will support significant future work.

## Acknowledgement

The research was supported in part the ARO (W911NF-16-1-0121) the NSF (IIS-1252835, IIS-1562364), and an Allen Distinguished Investigator Award. We would like to thank the reviewers for constructive feedback. Also thanks to Yotam Eshel and Noam Cohen for providing the Wikilink dataset. Special thanks to the members of UW NLP for helpful discussions and feedback.